\documentclass[conference]{IEEEtran}
\IEEEoverridecommandlockouts
\usepackage{cite}
\usepackage{amsmath,amssymb,amsfonts}
\usepackage{algorithmic}
\usepackage{graphicx}
\usepackage{textcomp}
\usepackage[table,xcdraw]{xcolor}
\usepackage{booktabs}
\usepackage{multirow}
\usepackage{graphicx}
\usepackage{subfigure}
\usepackage{url}

\def\BibTeX{{\rm B\kern-.05em{\sc i\kern-.025em b}\kern-.08em
    T\kern-.1667em\lower.7ex\hbox{E}\kern-.125emX}}
\begin{document}

\title{Instruction-Aligned Visual Attention for Mitigating Hallucinations in Large Vision-Language Models\\

\thanks{* contribute equally to this work and $\dagger$ 
 corresponding authors.}
}
\author{
\IEEEauthorblockN{Bin Li$^{*,1}$, Dehong Gao$^{*,2}$, Yeyuan Wang$^1$, Linbo Jin$^3$, Shanqing Yu$^{4,5}$, \\
Xiaoyan Cai$^{\dagger,1}$, Libin Yang$^{\dagger,2}$} \\
\IEEEauthorblockA{
\textit{$^1$School of Automation, Northwestern Polytechnical University, Xi’an, Shaanxi, China}\\
\textit{$^2$School of Cybersecurity, Northwestern Polytechnical University, Xi'an, Shaanxi, China}\\
\textit{$^3$Alibaba Group, Hangzhou, Zhejiang, China} \\
\textit{$^4$Zhejiang University of Technology, Hangzhou, Zhejiang, China} \\
\textit{$^5$Binjiang Institute of Artificial Intelligence, Hangzhou, Zhejiang, China} \\
\textit{\{libin0225, wangyeyuan\}}@mail.nwpu.edu.cn, 
\textit{\{dehong.gdh, xiaoyanc, libiny\}}@nwpu.edu.cn \\
\textit{yuyi.jlb}@alibaba-inc.com,
\textit{yushanqing}@zjut.edu.cn
} 
} 

\maketitle

\begin{abstract}
Despite the significant success of Large Vision-Language models(LVLMs), these models still suffer hallucinations when describing images, generating answers that include non-factual objects. 
It is reported that these models tend to over-focus on certain irrelevant image tokens that do not contain critical information for answering the question and distort the output. 
To address this, we propose an Instruction-Aligned Visual Attention(IAVA) approach,
which identifies irrelevant tokens by comparing changes in attention weights under two different instructions. 
By applying contrastive decoding, we dynamically adjust the logits generated from original image tokens and irrelevant image tokens, reducing the model’s over-attention to irrelevant information. 
The experimental results demonstrate that IAVA consistently outperforms existing decoding techniques on benchmarks such as MME, POPE, and TextVQA in mitigating object hallucinations. Our IAVA approach is available
online at \url{https://github.com/Lee-lab558/IAVA}.

\end{abstract}


\section{Introduction}
Large Vision Language Models (LVLMs) have recently demonstrated impressive performance in tasks such as image understanding and language generation~\cite{liu2023visual, InstructBLIP,wang2024cof}.
Combining visual and textual information, they have made significant breakthroughs on numerous benchmark tests\cite{zhuge2021kaleido} and are widely applied in fields like image captioning\cite{feng2024more,wang2024cogvlm} and visual question answering\cite{WU201721,ji2023masked,ma2024modula}. 
However, despite these advances, LVLMs still face the persistent challenge of \textbf{hallucination},
which refers to the phenomenon where the model generates plausible yet non-factual contents~\cite{liu2024survey}.
This issue not only affects the accuracy and reliability of LVLMs but can also lead to misleading or incorrect results in practical applications, which may undermine their dependability in critical areas such as healthcare\cite{luo2025chatgpt,mei2025adaptive,mei2024radchat,mei2024medical} and autonomous driving\cite{yurtsever2020survey,zhao2024autonomous,choudhary2024talk2bev}. 
Therefore, mitigating hallucinations and improving the model's ability to perceive reality accurately has become a crucial topic in current research.

\begin{figure}[ht] 
    \centering 
    \includegraphics[width=\columnwidth]{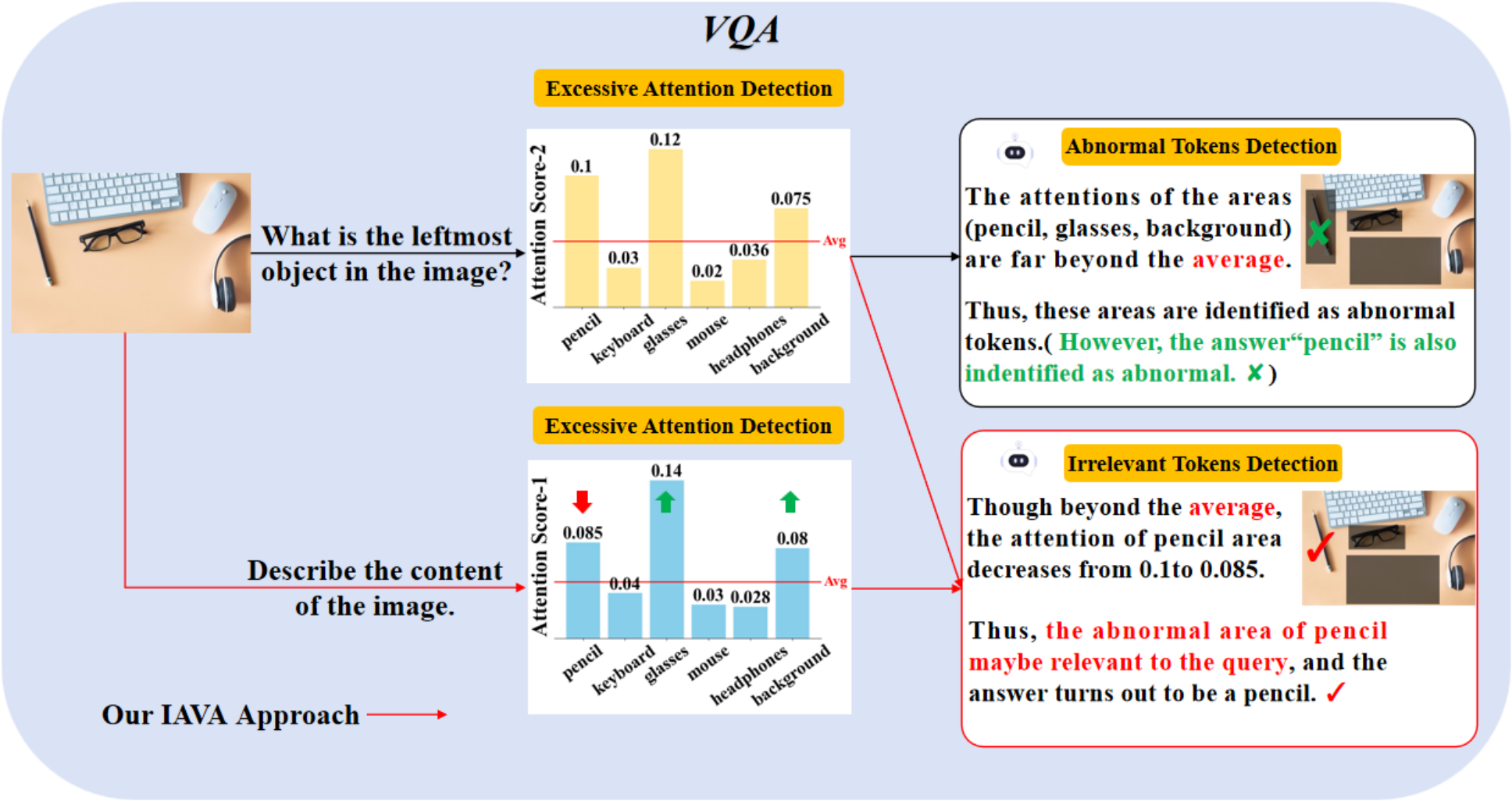}
    \caption{The irrelevant image tokens selection of our IAVA approach} 
    \label{fig_illustration} 
\end{figure}

Many studies have made progress in mitigating model hallucinations. 
These studies are categorized into three main directions: dataset dehallucination, modalities gap, and output correction~\cite{lan2024survey}.
Since dataset dehallucination and output correction approaches usually require external knowledge~\cite{liu2023mllms}\cite{hu2023ciem} or additional output processing~\cite{yin2023woodpecker}\cite{liu2025grounding}, this paper focuses on the modalities gap approaches, especially contrastive learning-based approaches.
These approaches leverage contrastive learning to exploit inconsistencies between positive and negative samples to minimize the modality gap between visual and genuine textual features~\cite{lan2024survey}. 
By comparing the inconsistencies, the final output probabilities of LVLMs can be adjusted to mitigate hallucinations. 
Existing studies have shown that models assign abnormally high attention to certain image tokens~\cite{woo2024dontmissforesttrees}\cite{gong-etal-2024-damro}, which means that a few image tokens occupy the majority of the attention. 
Notably, DAMRO\cite{gong-etal-2024-damro} further demonstrated that these abnormal image tokens are partially related to whether the model generates hallucinations. 
Therefore, existing methods use these abnormal image tokens as negative samples to amplify the model’s hallucination effects~\cite{woo2024dontmissforesttrees}\cite{gong-etal-2024-damro}. 
Indeed, models tend to overfocus on certain image tokens. As shown in Fig.~\ref{fig_illustration}, 
in the stage of excessive attention detection, only a few image tokens receive attention scores significantly higher than the average, while the remaining image tokens receive lower attention scores, with a more uniform distribution.
However, existing methods have not discussed whether these abnormal image tokens are relevant to the query target, but instead directly use these tokens to generate negative samples, which might constrain the LVLM performance.


To address this gap, we propose a learning-free contrastive decoding approach, called \textbf{I}nstruction-\textbf{A}ligned \textbf{V}isual \textbf{A}ttention (IAVA). 
The core of IAVA lies in distinguishing abnormal image tokens as negative samples while ensuring that tokens relevant to the query target are not mistakenly included, even if they receive excessively high attention. 
As shown in Fig.~\ref{fig_illustration}, the previous approach directly selects image tokens with excessively high attention, which may include tokens relevant to the query target. 
In contrast, our approach introduces an open-ended instruction to compute attention changes under two distinct instructions. 
This fine-grained information is then used to identify tokens that receive disproportionate attention but are irrelevant to the query. 
These tokens are ultimately chosen as negative samples. 
By leveraging contrastive decoding, our approach reduces the influence of irrelevant tokens on the response, adjusts the final output probabilities, mitigates hallucinations, and effectively enhances overall performance.  
The main contributions of this paper are as follows: 
\begin{itemize}
    \item We observe the phenomenon where LVLMs tend to over-focus on certain image tokens and propose using special instructions to amplify the role of prior knowledge, thereby identifying the regions the model is more inclined to focus on.
    \item We propose the IAVA approach, which focuses on identifying image tokens that occupy excessive attention but do not contain key information useful for answering the question, thus mitigating the impact of these irrelevant tokens and mitigating model hallucinations. 
    \item We conduct extensive experiments on MME, POPE, and TextVQA, comparing our IAVA with existing approaches and achieving significant improvement. On the MME dataset, we achieved the highest overall score, with improvements of approximately \(6.9\%\) and \(6.6\%\) over the baseline on the LLaVA and InstructBLIP models, respectively. Additionally, the results on the POPE dataset show that our approach achieved optimal performance across nearly all metrics. Finally, on the TextVQA dataset, We also improved the accuracy by approximately \(6.9\%\) and \(5.4\%\) compared to the baseline.
\end{itemize}

\section{Related works}
\subsection{Large Vision-Language Models}

Large vision-language models (LVLMs) have advanced significantly in integrating vision\cite{dosovitskiy2021an} and language\cite{chen2024general2specialized}, demonstrating strong performance in multimodal tasks like image captioning\cite{9739703,Kornblith_2023_ICCV,XU2023126287}, visual question answering (VQA), and image retrieval\cite{gao2020fashionbert,sun2024leveraging,feng2023vqa4cir}. Representative models (e.g., LLaVA-v1.5, InstructBLIP) achieve this by combining pre-trained language models\cite{vicuna2023,JMLR:v21:20-074}  with vision encoders\cite{dosovitskiy2021an,radford2021learning} for joint text-image understanding.


Recent research shows that attention mechanisms\cite{guo2022attention} in LVLMs determine which image parts are attended during inference, with attention maps visualizing the model's regional focus when processing questions. In this context, input instructions (e.g., captions, queries, structured prompts\cite{zhou2022learning,Shtedritski_2023_ICCV}) have gained increasing attention for their significant impact on the model's attention distribution across image regions.

\subsection{Hallucination in LVLMs}
With the evolution of Large Vision-Language Models (LVLMs), hallucination—defined as generating content misaligned with image inputs—has emerged as a critical challenge. This manifests when models perceive nonexistent objects (e.g., in empty scenes) or describe absent details, thereby compromising output accuracy and posing significant barriers to precision-sensitive applications.

Hallucination in LVLMs stems from multiple complex factors~\cite{liu2024survey,Hallucination_Lei_TOIS_24}. Firstly, constrained visual encoder resolution limits fine-grained detail capture (e.g., backgrounds, object counts) due to computational limitations\cite{dong2024internlm,huang2024hires}. Secondly, training data biases from noisy/inconsistent annotations in large-scale datasets promote hallucinations by reinforcing incorrect associations\cite{hu2023ciem,zhao2024looking}. Finally, dual priors exacerbate the issue: linguistic priors drive content supplementation through language patterns, while visual priors cause token over-focus, leading to misinterpretations\cite{wu2025lanp,lee2024vlind}.

\begin{figure*}[htbp]
    \centering
    \subfigure[Selection of Irrelevant Image Tokens]{
        \label{fig:2a}
        \includegraphics[width=0.9\linewidth]{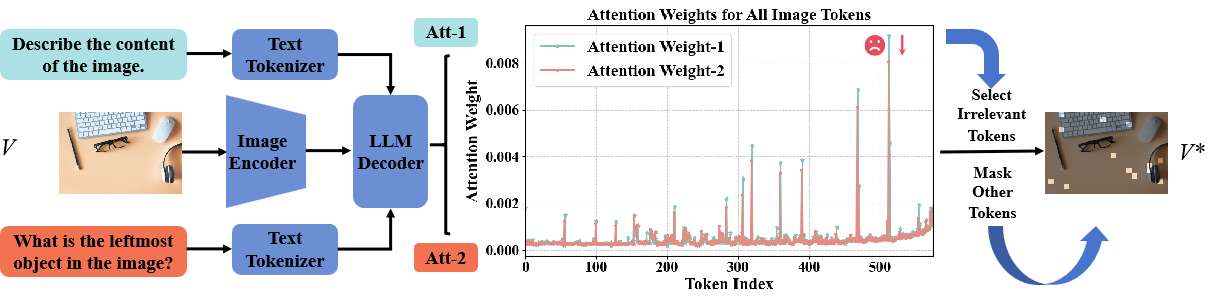}
    } \\
    \vspace{1em} 
    \subfigure[Contrastive Decoding Process]{
        \label{fig:2b}
        \includegraphics[width=0.8\linewidth]{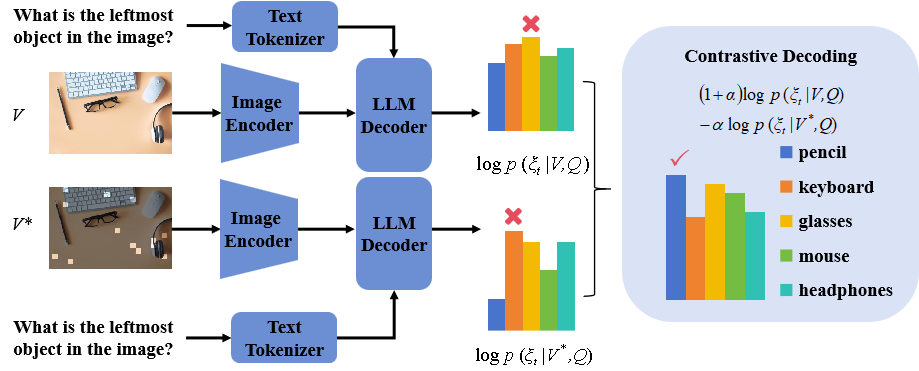}
    }
    \caption{Overview of the IAVA framework.
    (a) Identify which image tokens receive excessive attention but are irrelevant to the query by comparing the model’s attention scores under two different instructions.
    (b) Use the irrelevant image tokens obtained in (a) and perform contrastive decoding with the original image to mitigate the influence of these tokens.}
    \label{fig:2}
\end{figure*}

\subsection{Contrastive Decoding}
Contrastive decoding techniques have recently made significant strides in mitigating hallucinations in LVLMs. 
The core idea of contrastive decoding is to leverage difference between positive and negative samples to adjust model output probabilities\cite{wang2024enhancing}, thereby reducing the model’s focus on irrelevant information. 
This approach has been validated by numerous studies for its effectiveness in reducing hallucinations\cite{Leng_2024_CVPR}\cite{DBLP:journals/corr/abs-2403-18715}\cite{Favero_2024_CVPR}.

 Various methods have been explored for selecting negative samples. For example,VCD\cite{Leng_2024_CVPR} uses images with added Gaussian noise as negative samples, where the noise introduces greater visual uncertainty. In contrast, M3ID\cite{Favero_2024_CVPR} selects pure text inputs as negative samples. Both methods intensify the model's reliance on linguistic priors, exacerbating hallucination. In our experiments, we also compare M3ID and VCD with our IAVA approach to better demonstrate the effectiveness and feasibility of our approach.

\section{Methodology}
\subsection{Overview}\label{AA}
The overview of the proposed IAVA approach is illustrated in Fig.~\ref{fig:2},  
of which the core lies in effectively identifying image tokens irrelevant to the query text. 
To this end, we introduce two instructions to enforce the LVLMs to perform two separate attention calculations before generating a response.
The magnitude and variation of attention scores between these two calculations serve as an irrelevance metric.
After identifying the irrelevant image tokens, masking is applied to the original image, filtering to retain only the irrelevant tokens while removing the others. 
With contrastive decoding, the output probabilities of LVLMs are adjusted, thereby mitigating hallucination effects.

\subsection{Irrelevant Image Tokens}
As shown in Fig.~\ref{fig:2}(a), when selecting image tokens that are irrelevant to the query, we prepare two different instructions. 
The first instruction uses a general instruction like ``Describe the content of the image'', while the second uses the original query as input. 
In the absence of a specific query, the first general instruction guides the model in broadly describing the image content, and encouraging it to focus on general information rather than the specific target of the query. 
These tokens typically do not contain information useful for answering a specific question but are instead relevant to the general image content. 
In this way, we can identify areas where the model might over-focus when describing the image and further assess whether these areas offer critical cues in actual query tasks.  
Subsequently, with the second instruction (e.g., ``What is the leftmost object in the image?''), LVLMs tend to pay more attention to the provided query. 
By comparing the attention distributions under these two instructions, we can better understand attention variations, identifying which image tokens are overly focused on when there is no specific query but are irrelevant to the answer when the query is provided. 
Specifically, the attention distribution under the first instruction reveals potential hallucination regions that the model might focus on in the ``free description'' mode, while the attention distribution under the second instruction highlights the relevant regions for the actual query. 
By comparing the differences between the two attention distributions, we can accurately select those irrelevant image tokens, generate negative samples for them, and reduce the influence of these areas in the generation of responses during the subsequent contrastive decoding process to mitigate the impact of hallucination phenomena. 

\begin{figure*}[htbp]
    \centering
    \subfigure[InstructBLIP]{
        \label{fig:3a}
        \includegraphics[width=0.48\linewidth]{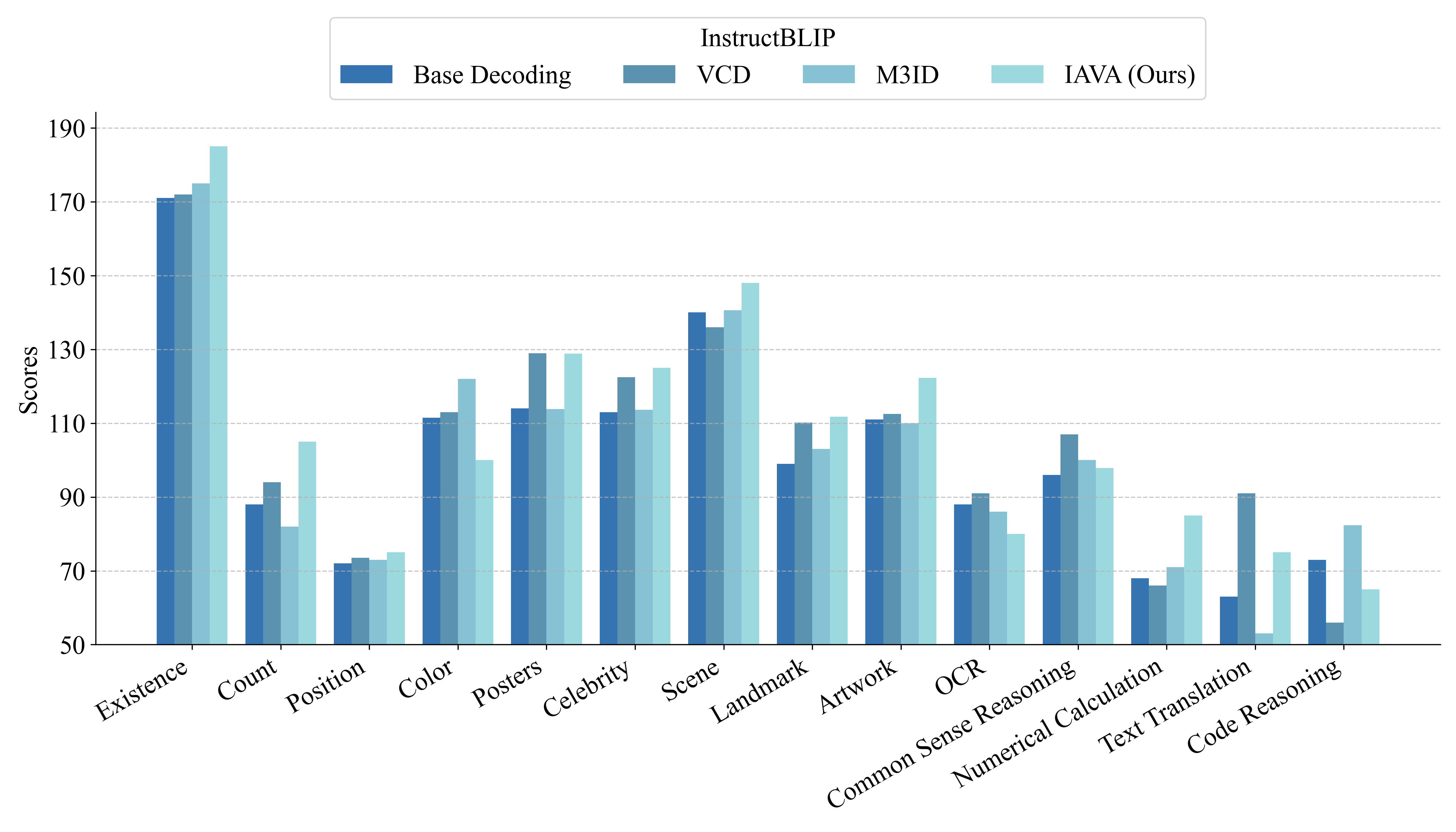}
    }
    \subfigure[LLaVA-1.5]{
        \label{fig:3b}
        \includegraphics[width=0.48\linewidth]{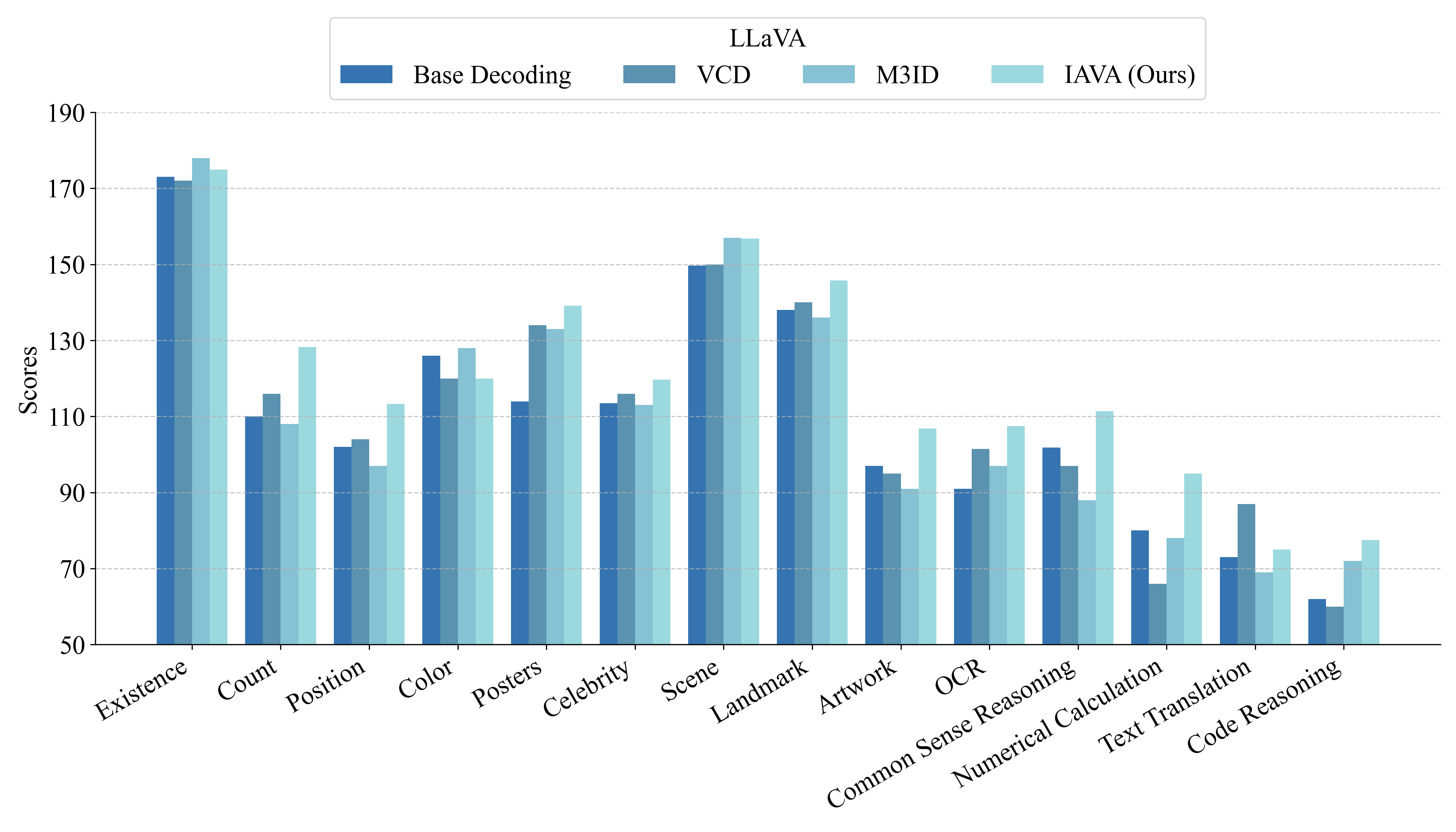}
    }
    \caption{Result comparison on MME.
The IAVA approach achieved the highest overall score on both models, with the best performance on 10 sub-tasks when using LLaVA, and the best performance on 8 sub-tasks when using InstructBLIP.}
    \label{fig:3}
\end{figure*}

We use the following two conditions to measure which image tokens are irrelevant: (1) the attention score of the image token decreases with the two instructions; 
(2) the image token receives a high attention score when the general instruction is input. 
For those image tokens that satisfy both of these conditions, it indicates that they occupy excessive attention but do not contain key information for answering the question, and thus, they are labeled as irrelevant image tokens. 
We first define the attention scores for the general instruction and the actual query as \(att_1[idx]\) and \(att_2[idx]\), respectively, and compute the mean \(\mu\) and standard deviation \(\sigma\) of the former as shown in \eqref{eq:1} and \eqref{eq:2}. 

\begin{equation}\label{eq:1}
\mu=\frac{1}{n+1} \sum_{idx=0}^{n} att_1[idx]
\end{equation}
\begin{equation}\label{eq:2}
\sigma=\sqrt{\frac{1}{n+1} \sum_{idx=0}^{n} (att_1[idx] - \mu)^2}
\end{equation}
Then, we calculate the attention change for the image token as shown in \eqref{eq:3} to obtain \({\Delta}att[idx]\), and sort it to get \({\Delta}att^{sort}[i]\), where idx is the original image token index, and \(i\) is an adjustable parameter. Finally, the selection rule for irrelevant tokens ($it$) is given in \eqref{eq:4}, where \(\lambda\) is also an adjustable parameter, which, together with parameter \(i\), controls the number of irrelevant tokens selected.  


\begin{equation}\label{eq:3}
\begin{aligned}
\Delta att[idx] = att_{2}[idx] - att_{1}[idx] 
\end{aligned}
\end{equation}
\begin{equation}\label{eq:4}
\begin{aligned}
it[idx] &= \{j | \; \Delta att[j] < 0, \\
& \qquad\quad {\Delta}att[j] < {\Delta}att^{sort}[i], \\ 
& \qquad\quad att_1[j] > \mu + \lambda\sigma\}
\end{aligned}
\end{equation}

\subsection{Contrastive Decoding Process}
The irrelevant image tokens are identified in the previous section.
With the irrelevant image token, contrastive decoding is introduced to mitigate the hallucination phenomenon.
We mask the original image \(V\), keeping only the irrelevant tokens while masking the others, resulting in a masked image \(V^*\) which is treated as a negative sample. 
Contrastive decoding is typically performed during the sampling process in LLMs, where the next predicted token is determined on the basis of the probability distribution in the logit space. 
As shown in Fig.~\ref{fig:2}(b), for the same query text \(Q\), we provide both the original image \(V\) and the masked image \(V^*\) as input, then predict the next token \(\xi_t\) using the \eqref{eq:4}.
\begin{equation}\label{eq:5}
\begin{aligned}
\xi_t=Softmax\{ & (1+\alpha)log_{\theta}p(\xi_t|\xi<t,V,Q)-\\ 
                & \alpha log_{\theta}p(\xi_t|\xi<t,V^*,Q)\}
\end{aligned}
\end{equation}

In \eqref{eq:4}, we compute the original output probability distribution based on \(V\) and \(Q\), and the biased output probability distribution based on \(V^*\) and \(Q\). 
Using the negative sample \(V^*\), the model tends to generate hallucinations, increasing the probability of incorrect output. 
Then we adjust the final output through the contrastive strength parameter \(\alpha\), effectively reducing the likelihood of hallucinations in LVLMs and enhancing the model's performance.  

\section{Experiments}
\subsection{Experimental Settings}

\subsubsection{Baseline}
Our IAVA approach is evaluated on LLaVA-v1.5 and InstructBLIP, and compared with two contrastive decoding methods: VCD\cite{Leng_2024_CVPR} (adding noise to images) and M3ID\cite{Favero_2024_CVPR} (using pure text input). These methods, including IAVA, require no additional training or fine-tuning. To demonstrate performance improvements, all enhanced decoding methods are benchmarked against the original models under standard base decoding.

\subsubsection{Implementation Details}
The parameter settings during the experiments are as follows. 
For both models, the random seed is fixed at 42, and the parameter \(\alpha\) in the contrastive decoding described in \eqref{eq:4} is set to 1. 
Equation \eqref{eq:3} explains the conditions for selecting irrelevant image tokens. 
For the LLaVA model, which has 576 image tokens, \(i\) is set to 292 and \(\lambda\) to 0. 
For the InstructBLIP model, which has 32 image tokens, \(i\) is set to 16 and \(\lambda\) to -0.1.

\subsection{Benchmarks and Experimental Results}
Our approach is experimented with three datasets: MME, POPE, and TextVQA. 
MME consists of 14 sub-tasks, and evaluating these tasks provides a comprehensive assessment of the model’s perceptual and cognitive abilities. 
POPE translates the hallucination evaluation into a series of binary questions about whether objects are present in the image. 
The TextVQA dataset is specifically designed for Visual Question-Answering (VQA) tasks, with a particular emphasis on understanding textual information within the image.  

\begin{table}[b]
\caption{Result comparison on POPE}
\resizebox{\columnwidth}{!}{%
\begin{tabular}{cccccccccc}
\hline
                        &                          & \multicolumn{4}{c}{InstructBLIP} & \multicolumn{4}{c}{LLaVA 1.5}                                                 \\ \cline{3-10} 
\multirow{-2}{*}{Setup} & \multirow{-2}{*}{Method} & Acc.   & Prec.  & Rec.   & F1    & Acc.  & Prec. & Rec.                          & F1                            \\ \hline
                        & base                     & 82.27  & 82.84  & 81.40  & 82.11 & 84.47 & 83.35 & 86.13                         & 84.72                         \\
                        & VCD                      & 83.37  & 83.39  & 82.60  & 83.24 & 84.80 & 83.00 & 87.53                         & 85.20                         \\
                        & M3ID                     & 84.37  & 84.62  & 84.00  & 84.31 & 86.00 & 85.11 & 87.27                         & 86.18                         \\
\multirow{-4}{*}{Random} &
  IAVA &
  \cellcolor[HTML]{FFFC9E}86.91 &
  \cellcolor[HTML]{FFFC9E}88.45 &
  \cellcolor[HTML]{FFFC9E}85.80 &
  \cellcolor[HTML]{FFFC9E}87.11 &
  \cellcolor[HTML]{FFFC9E}88.04 &
  \cellcolor[HTML]{FFFC9E}88.87 &
  \cellcolor[HTML]{FFFC9E}87.80 &
  \cellcolor[HTML]{FFFC9E}88.33 \\ \hline
                        & base                     & 77.77  & 74.81  & 83.73  & 79.02 & 82.23 & 79.72 & 86.47                         & 82.95                         \\
                        & VCD                      & 78.00  & 75.12  & 83.73  & 79.19 & 82.27 & 79.19 & 87.53                         & 83.15                         \\
                        & M3ID                     & 77.30  & 74.10  & 83.93  & 78.71 & 82.83 & 79.62 & \cellcolor[HTML]{FFFC9E}88.27 & 83.72                         \\
\multirow{-4}{*}{Popular} &
  IAVA &
  \cellcolor[HTML]{FFFC9E}80.97 &
  \cellcolor[HTML]{FFFC9E}77.40 &
  \cellcolor[HTML]{FFFC9E}87.47 &
  \cellcolor[HTML]{FFFC9E}82.13 &
  \cellcolor[HTML]{FFFC9E}84.77 &
  \cellcolor[HTML]{FFFC9E}82.70 &
  87.93 &
  \cellcolor[HTML]{FFFC9E}85.23 \\ \hline
                        & base                     & 73.13  & 69.41  & 82.60  & 75.46 & 77.10 & 72.57 & 87.13                         & 79.19                         \\
                        & VCD                      & 75.87  & 72.85  & 82.47  & 77.36 & 76.10 & 71.50 & 86.80                         & 78.41                         \\
                        & M3ID                     & 76.03  & 72.47  & 83.93  & 77.79 & 77.70 & 73.23 & 87.33                         & \cellcolor[HTML]{FFFFFF}79.66 \\
\multirow{-4}{*}{Adversarial} &
  IAVA &
  \cellcolor[HTML]{FFFC9E}77.93 &
  \cellcolor[HTML]{FFFC9E}73.57 &
  \cellcolor[HTML]{FFFC9E}87.20 &
  \cellcolor[HTML]{FFFC9E}79.80 &
  \cellcolor[HTML]{FFFC9E}78.17 &
  \cellcolor[HTML]{FFFC9E}73.67 &
  \cellcolor[HTML]{FFFC9E}87.67 &
  \cellcolor[HTML]{FFFC9E}80.06 \\ \hline
\end{tabular}%
}
\label{table:1}
\end{table}

As shown in Fig.~\ref{fig:3}, the experimental results demonstrate that our IAVA approach achieves the highest overall score on the MME dataset. 
Specifically, when using the InstructBLIP model, IAVA outperforms other approaches in 8 out of the 14 sub-tasks, while with the LLaVA model, it achieves the best performance in 10 sub-tasks. 
These results highlight the adaptability and robustness of our approach across different tasks and model architectures. 
Furthermore, Table~\ref{table:1} presents the experimental results on the POPE dataset, where IAVA consistently achieves superior performance across almost all metrics, demonstrating its effectiveness in mitigating hallucinations and improving response accuracy. 
The results validate the ability of our method to align visual attention more effectively, making it a reliable solution for enhancing LVLMs in diverse evaluation scenarios. 
Meanwhile, Table~\ref{tabel:2} presents the experimental results of our method on the TextVQA dataset. 
Compared to the reproduced results of base decoding, our method achieves approximately \(7\%\) and \(5\%\) improvements on the two models, respectively. 
It can also be observed that the M3ID approach achieves higher accuracy than ours on the LLaVA model, but our IAVA approach is not far behind. 
This shows that our approach remains effective even in scenarios that emphasize the understanding of textual information within images.  
\begin{table}[!]
\caption{Result comparison on TextVQA}
\centering
\resizebox{0.55\columnwidth}{!}{
\begin{tabular}{ccc}
\hline
                         & \multicolumn{2}{c}{Accuracy}                                      \\ \cline{2-3} 
\multirow{-2}{*}{Method} & LLaVA 1.5                       & InstructBLIP                    \\ \hline
base                     & 43.48\%                         & 23.48\%                         \\
VCD                      & 47.32\%                         & 28.22\%                         \\
M3ID                     & \cellcolor[HTML]{FFFC9E}51.56\% & 27.23\%                         \\
IAVA                     & 50.35\%                         & \cellcolor[HTML]{FFFC9E}28.84\% \\ \hline
\end{tabular}%
}
\label{tabel:2}
\end{table}
\begin{figure}[b] 
    \centering \includegraphics[width=0.8\columnwidth]{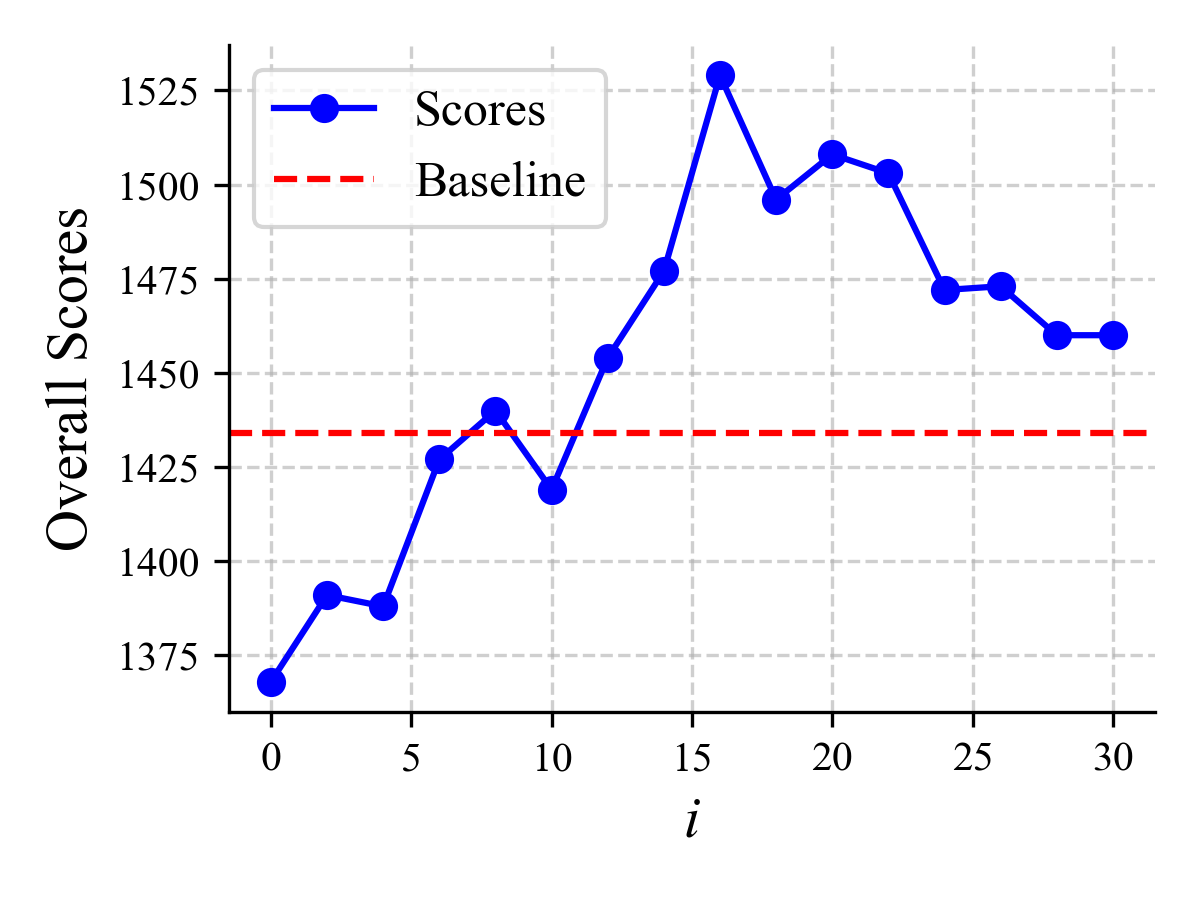}
    \caption{Result Variation on MME with Parameter \(i\)} 
    \label{fig4} 
\end{figure}

\subsection{Ablation Study}
Since certain image tokens are masked during the generation of negative samples, the visual information contained in the original image is inevitably damaged. 
To better understand the impact of the IAVA approach on model performance, we adjust the parameter \(i\) in \eqref{eq:3} to control the number of masked tokens, and test on the MME dataset. 
The overall score is recorded against different values of \(i\). 
In this ablation study, InstructBLIP is selected as the experimental model because it uses only 32 image tokens, allowing a comprehensive observation of performance variations with respect to \(i\). 
The final results are presented in Fig.~\ref{fig4}. 
Based on the variation of the overall score of InstructBLIP on MME as \(i\) changes, the following analyses can be made:  

According to the conditions measuring which image tokens are irrelevant given in \eqref{eq:3}, the number of irrelevant image tokens is positively correlated with the parameter \(i\). 
Specifically, as \(i\) decreases, the number of irrelevant tokens decreases as well. 
However, since the masked images used as negative samples retain only these irrelevant tokens, a small \(i\) value results in images that lose substantial visual information. 
In such cases, the model struggles to extract meaningful information from the image when answering questions, leading to highly unstable performance.

When \(i\) is set to a moderate value, the masked images primarily retain visual information irrelevant to the query. 
These images are ideal as negative samples to effectively disrupt the model’s visual processing. At this stage, the contrast between positive and negative samples achieves its maximum impact, leading to optimal performance.
As \(i\) continues to increase, the masked regions progressively shrink until they stabilize. 
During this process, some image tokens that were previously masked and potentially relevant to the query may instead be retained due to the larger \(i\). 
This change reduces the distance between positive and negative samples, weakens the contrastive effect, and consequently lowers the model's performance.

The above conclusions demonstrate the rationality of the parameter settings during the experiments. 
Additionally, the results of the ablation study further validate that our IAVA approach can effectively identify image tokens irrelevant to the query and use them as negative samples, maximizing the distance between positive and negative samples. 
This ensures the optimal effectiveness of contrastive decoding, leading to a significant improvement in model performance.

\section{Conclusion}
This paper proposes a novel IAVA approach, which uses two different instructions to identify image tokens that are irrelevant to the query text and mitigates their impact during the decoding stage, aligning the model's visual attention with the query text. 
This effectively mitigates the hallucination phenomenon in LVLMs. 
Our approach adjusts the model's attention distribution through contrastive decoding, reducing the interference of irrelevant image tokens in generating answers. 
Experiments on datasets such as MME, POPE, and TextVQA show that IAVA outperforms existing contrastive decoding methods (e.g., VCD and M3ID) in mitigating hallucinations and improving model performance. 
Our research provides new insights into enhancing the reliability and accuracy of LVLMs, laying the foundation for future improvements.

\section{Acknowledgements}
This work was supported in part by the National Natural Science Foundation of China under Grants 62372380 and U22B2036, Key Research and Development Program of Shaanxi Province under Grants 2024GX-ZDCYL-01-05, the Natural Science Basic Research Program of Shaanxi under Grants 2024JC-YBMS-513, Key Research and Development Program of Zhejiang Province under Grants 2024C01025, and and Key Research and Development Program of Hangzhou under Grants 2024SZD1A23.

\bibliographystyle{IEEEbib}
\bibliography{icme2025references}

\end{document}